\documentclass[10pt,twocolumn,letterpaper]{article}

\usepackage{iccv}              

\definecolor{mygreen}{RGB}{84, 130, 53}

\usepackage{multirow}
\usepackage{arydshln}
\usepackage{colortbl}
\definecolor{Gray}{gray}{0.95}
\newcommand{\gb}[1]{{\cellcolor{gray!18}{#1}}}
\usepackage{booktabs}
\definecolor{iccvblue}{rgb}{0.21,0.49,0.74}
\usepackage[pagebackref,breaklinks,colorlinks,allcolors=iccvblue]{hyperref}

\def\paperID{8337} 
\def\confName{ICCV}
\def\confYear{2025}

\title{Lay2Story: Extending Diffusion Transformers for Layout-Togglable Story Generation}

\author{
Ao Ma\textsuperscript{*,\dag}, Jiasong Feng\textsuperscript{*}, Ke Cao\textsuperscript{*}, Jing Wang\textsuperscript{*}, Yun Wang, Quanwei Zhang, Zhanjie Zhang\textsuperscript{\ddag} \\
JD.com, Inc., Beijing, China \\
{\tt\small maao.8@jd.com, zhangzhanj@126.com}}

\begin{document}

\twocolumn[{%
\renewcommand\twocolumn[1][]{#1}%
\maketitle
\vspace{-9mm}
\begin{center}
    \centering
    \includegraphics[width=\linewidth]{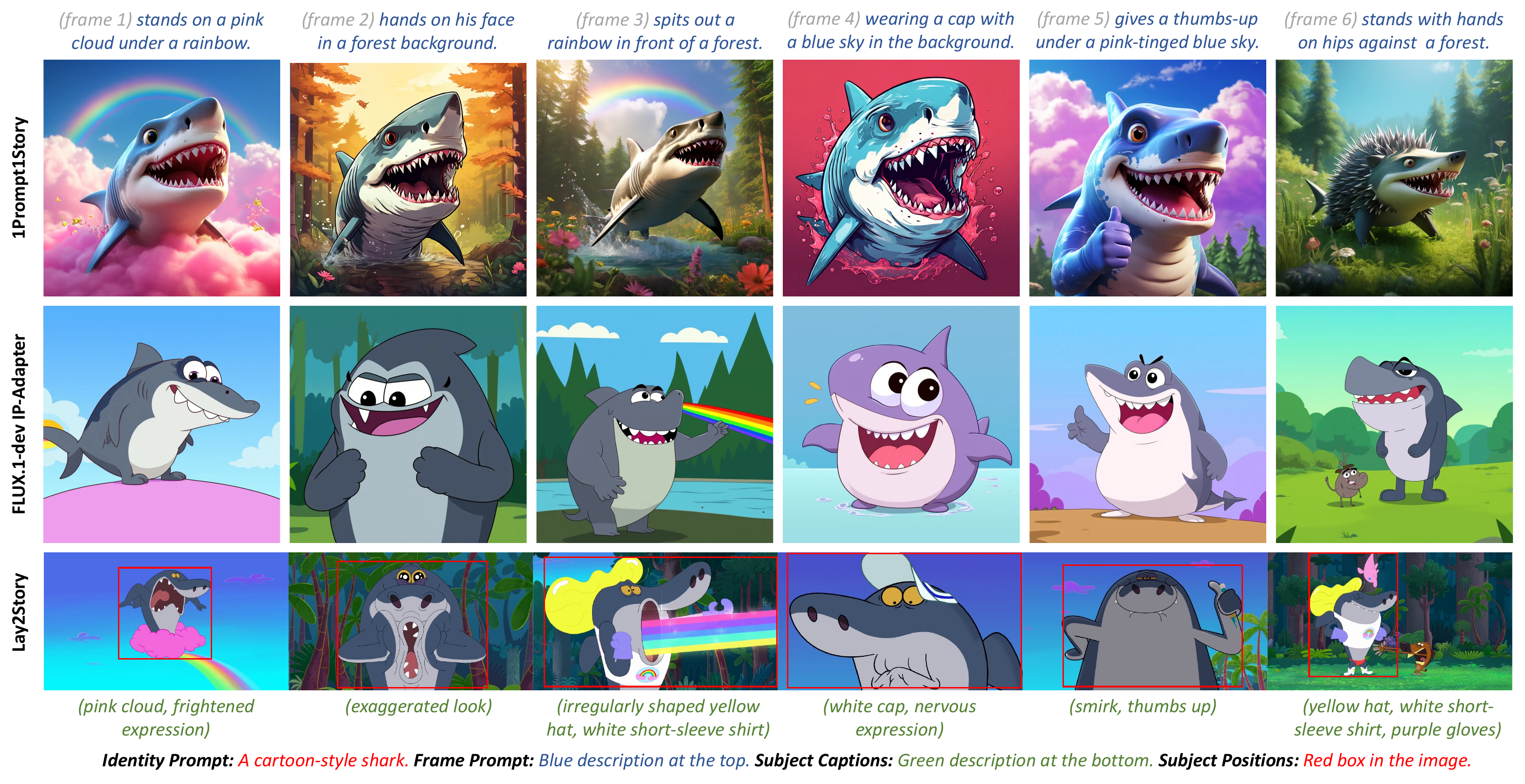}
    \vspace{-6mm}
    \captionof{figure}{
    \textbf{Comparison with SOTA methods.} Lay2Story offers finer subject control in a 6-frame story with subject captions (\textcolor{mygreen}{green description}) and positions (\textcolor{red}{red box}), unlike the contrast method, which uses only identity and frame prompts. It outperforms 1Prompt1Story and FLUX.1-dev IP-Adapter in image consistency, generating images at 720 $\times$ 1280 resolution, while others use 1024 $\times$ 1024.
    } \vspace{1mm}
    \label{figure1}
\end{center}%
\vspace{-2mm}
}]

\begingroup
\renewcommand\thefootnote{}%
\footnotetext{%
\textsuperscript{*} Equal contribution.\hspace{1em}%
\textsuperscript{\dag} Project leader.\hspace{1em}%
\textsuperscript{\ddag} Corresponding author.%
}
\endgroup

\begin{abstract}

Storytelling tasks involving generating consistent subjects have gained significant attention recently. However, existing methods, whether training-free or training-based, continue to face challenges in maintaining subject consistency due to the lack of fine-grained guidance and inter-frame interaction. Additionally, the scarcity of high-quality data in this field makes it difficult to precisely control storytelling tasks, including the subject's position, appearance, clothing, expression, and posture, thereby hindering further advancements. In this paper, we demonstrate that layout conditions, such as the subject's position and detailed attributes, effectively facilitate fine-grained interactions between frames. This not only strengthens the consistency of the generated frame sequence but also allows for precise control over the subject’s position, appearance, and other key details. Building on this, we introduce an advanced storytelling task: Layout-Togglable Storytelling, which enables precise subject control by incorporating layout conditions. To address the lack of high-quality datasets with layout annotations for this task, we develop Lay2Story-1M, which contains over 1 million 720p and higher-resolution images, processed from approximately 11,300 hours of cartoon videos. Building on Lay2Story-1M, we create Lay2Story-Bench, a benchmark with 3,000 prompts designed to evaluate the performance of different methods on this task. Furthermore, we propose Lay2Story, a robust framework based on the Diffusion Transformers (DiTs) architecture for Layout-Togglable Storytelling tasks. Through both qualitative and quantitative experiments, we find that our method outperforms the previous state-of-the-art (SOTA) techniques, achieving the best results in terms of consistency, semantic correlation, and aesthetic quality.
\end{abstract}
\vspace{-6mm}
\section{Introduction}
\label{Introduction}
\vspace{-1mm}

The storytelling task focuses on generating consistent images from text prompts and has gained interest with the rise of diffusion models ~\cite{rombach2022high,ramesh2022hierarchical,liu2025bridge,wang2025wisa,ma2024hico,zhang2024artbank,zhang2025vectorsketcher,
zhang2025spast,ling2025ragar}. The current work ~\cite{singhstorybooth, liu2025one, mao2024story, tewel2024training, he2025anystory, li2013story, alhussain2021automatic, yang2024seed, ma2024storynizor, zhou2024storymaker, liu2024intelligent, zhou2025storydiffusion} follows two main approaches: training-free and training-based methods. Training-free methods ~\cite{singhstorybooth, liu2025one, mao2024story, tewel2024training, zhou2025storydiffusion,shao2025eventvad} typically modify the cross-frame self-attention of the original text-to-image (T2I) model to generate consistent frame sequences. These methods are versatile, enabling application across various domains and diffusion model configurations. Training-based methods ~\cite{he2025anystory,yang2024seed, ma2024storynizor, zhou2024storymaker, liu2024intelligent, flux-ipa,wang2025dualnet,
zhang2025lgast,wang2025adstereo,lu2024recent} ensure subject consistency by learning from sequential frames, capturing complex visual concepts from the training set.

However, both training-free and training-based methods lack fine-grained guidance and inter-frame interaction, making it challenging to maintain subject consistency. Additionally, existing methods struggle to precisely control the subject within the frame, including position, appearance, clothing, expression, and posture, due to the absence of large-scale datasets with subject refinement annotations, hindering further advancements in this field.

In this paper, we first explore the potential of using layout conditions for precise control over the subject. As shown in Fig.~\ref{figure1}, we present our method with layout conditions alongside the current state-of-the-art (SOTA) training-free method, 1Prompt1Story ~\cite{liu2025one}, and the training-based method, FLUX.1-dev IP-Adapter ~\cite{flux-ipa}, providing a qualitative comparison. The results show that, compared to existing methods, our approach effectively guides fine-grained interactions between frames using layout conditions. This not only enhances the consistency of the generated frame sequence but also enables precise control over the subject's position, appearance, and other details. Building on this, we introduce a novel advanced storytelling task: Layout-Togglable Storytelling, which enables precise subject control by incorporating layout conditions. To address the scarcity of large-scale, high-quality datasets with layout annotations for this task, we present Lay2Story-1M, a dataset consisting of over 1 million images at 720p resolution or higher, complete with detailed subject annotations. Building upon Lay2Story-1M, we also introduce Lay2Story-Bench, a benchmark containing 3,000 prompts and corresponding high-quality images to evaluate the performance of our model and comparison methods.

To incorporate layout conditions and enable more refined subject storytelling, we propose Lay2Story, a training-based method designed for Layout-Togglable Storytelling tasks. We select PixArt-$\alpha$ ~\cite{chen2023pixarta} as our base model, a T2I model built on the Diffusion Transformers (DiTs) ~\cite{peebles2023scalable}. Inspired by some methods ~\cite{zavadski2024controlnet, cao2025relactrl,zhao2023uni,zhang2024towards,zhang2025u,lu2025uni,bi2025customttt}, our model comprises two main branches: the global branch and the subject branch. The global branch takes noise latent as input, guided by global captions, and focuses on generating the overall image content. The subject branch takes as input the noise latent, subject mask, and latent vector of a reference image, guided by the subject captions, focusing on maintaining subject consistency and generating subject position and detailed information. In the experiments, we compare our method with several SOTA approaches, including BLIP-Diffusion ~\cite{li2023blip}, StoryGen ~\cite{liu2024intelligent}, ConsiStory \cite{tewel2024training}, StoryDiffusion ~\cite{zhou2025storydiffusion}, 1Prompt1Story ~\cite{liu2025one}, and FLUX.1-dev IP-Adapter ~\cite{flux-ipa}, among others. Both qualitative and quantitative evaluations confirm the effectiveness of our method. In summary, the main contributions of this paper are as follows:
\begin{itemize}
\item We propose an advanced version of the storytelling task: \textbf{Layout-Togglable Storytelling}, which enables precise subject control by incorporating layout conditions. It ensures subject consistency while offering detailed control over the subject, including its position, appearance, clothing, expression, posture, and other relevant details.
\item We create the \textbf{Lay2Story-1M} dataset, the largest storytelling dataset to the best of our knowledge, consisting of over 1 million 720p and higher resolution images with detailed subject annotations. Building on this, we present Lay2Story-Bench, a benchmark consisting of 3,000 prompts and corresponding high-quality images, designed to evaluate our model and related methods.
\item We introduce \textbf{Lay2Story}, a training-based approach built on the DiTs architecture, designed for Layout-Togglable Storytelling tasks. Through extensive comparisons with existing storytelling methods, we confirm that Lay2Story outperforms relevant approaches in terms of consistency, semantic correlation, and aesthetic quality.
\end{itemize}

\vspace{-2mm}
\section{Related Work}
\label{sec:Related_Work}
\vspace{-1mm}

Please refer to the \textcolor{blue}{\textbf{Supplementary Material A}}.

\vspace{-2mm}
\section{Lay2Story-1M}

\indent\textbf{Lay2Story-1M} is a dataset specifically designed for the Layout-Togglable Storytelling task. It consists of approximately 200,000 frame sequences, each containing 4 to 6 images, all featuring the same subject and a resolution of at least 720p. In total, the dataset contains around 1 million images. Each image is annotated with a global caption, following the format ``identity prompt + frame prompt'', similar to other storytelling approaches ~\cite{liu2025one, singhstorybooth, tewel2024training, ma2024storynizor, zhou2025storydiffusion}. To achieve more precise control over subject generation, we annotate the layout conditions, including subject positions and descriptive captions. Some examples of Lay2Story-1M are shown in the \textcolor{blue}{\textbf{Supplementary Material B.1.}}

\begin{table}[h]
\caption{\textbf{Comparison of existing storytelling datasets.} Compared to existing datasets like StorySalon, StoryDB, and StoryStream, our Lay2Story-1M features a larger image collection, higher resolution, and more detailed subject annotations.}
\resizebox{\linewidth}{!}{
\begin{tabular}{cccc}
\hline
Datasets & Images Num & Resolution & \multicolumn{1}{c}{Detailed Annotations} \\
         &            &            & \multicolumn{1}{c}{of Subjects} \\ \hline
StorySalon ~\cite{liu2024intelligent} & 159,778 & 432 $\times$ 803 & no \\
StoryDB ~\cite{ma2024storynizor} & 100,000 & 512 $\times$ 512 & no \\
StoryStream ~\cite{yang2024seed} & 257,850 & 480 $\times$ 854 & no  \\
\textbf{Lay2Story-1M} & 1,020,600 & 720 $\times$ 1080 & yes \\ \hline
\end{tabular}}
\label{table1}
\vspace{-2mm}
\end{table}

To simplify the task and reduce workload, we focus on \textbf{cartoon} scene data, labeling only \textbf{the most prominent} subject character in each frame, even when multiple subjects appear. We compare Lay2Story-1M with several storytelling datasets presented in existing papers ~\cite{liu2024intelligent, ma2024storynizor, yang2024seed}, as shown in Tab.~\ref{table1}. Lay2Story-1M is the largest high-resolution storytelling dataset known to us so far, with detailed subject tagging. We believe that it can also be used for tasks beyond storytelling, such as high-quality cartoon image generation and layout-to-image generation. Our data collection and filtering process is described in Sec.~\ref{Data Collection and Filtering}. Next, we present our frame sequence data construction pipeline in Sec.~\ref{Frame Sequence Construction}. Finally, in Sec.~\ref{Lay2Story-Bench}, we introduce our test set, Lay2Story-Bench, based on Lay2Story-1M.

\subsection{Data Collection and Filtering}
\label{Data Collection and Filtering}
\textbf{Video Collection.} Considering the inherent consistency of subjects within video data, we select video as our primary source of data. Following \cite{wang2023internvid, chen2024panda, ju2025miradata, wang2023videofactory, wang2024qihoo, che2024gamegen, feng2024fancyvideo, nan2024openvid}, we primarily collect and download video data from the Internet, with our video sources covering the following three types.
(1) \textbf{PBS Kids and Khan Academy.} We collect approximately 12,000 cartoon videos from platforms and organizations that offer copyright-friendly content, such as PBS Kids \footnote{\url{https://pbskids.org/}} and Khan Academy \footnote{\url{https://www.khanacademy.org/}}, which support nonprofit initiatives focused on education and academia.
(2) \textbf{Internet Archive.} We collect about 8,000 videos from the Internet Archive \footnote{\url{https://web.archive.org/}} that are in the public domain and no longer under copyright for use in our academic projects.
(3) \textbf{YouTube Videos.} We collect approximately 20,000 high-quality cartoon videos from YouTube \footnote{\url{https://www.youtube.com/}} and implement the following measures to mitigate potential copyright issues. First, the data used in our study are sourced from public YouTube channels and do not involve any exclusive or private data sources. Second, in line with previous works \cite{wang2023internvid, chen2024panda, ju2025miradata, jiang2023res, che2024gamegen,he2025plangen}, our data-sharing strategy involves providing only the YouTube video IDs necessary to download the corresponding content, along with our code to process the data, rather than sharing raw data. Finally, our data collection and publication practices fully comply with YouTube's Data Privacy Policy and Fair Use Policy, and the released data is intended solely for research purposes.
\begin{figure}
  \centering
    \includegraphics[width=1.0\linewidth]{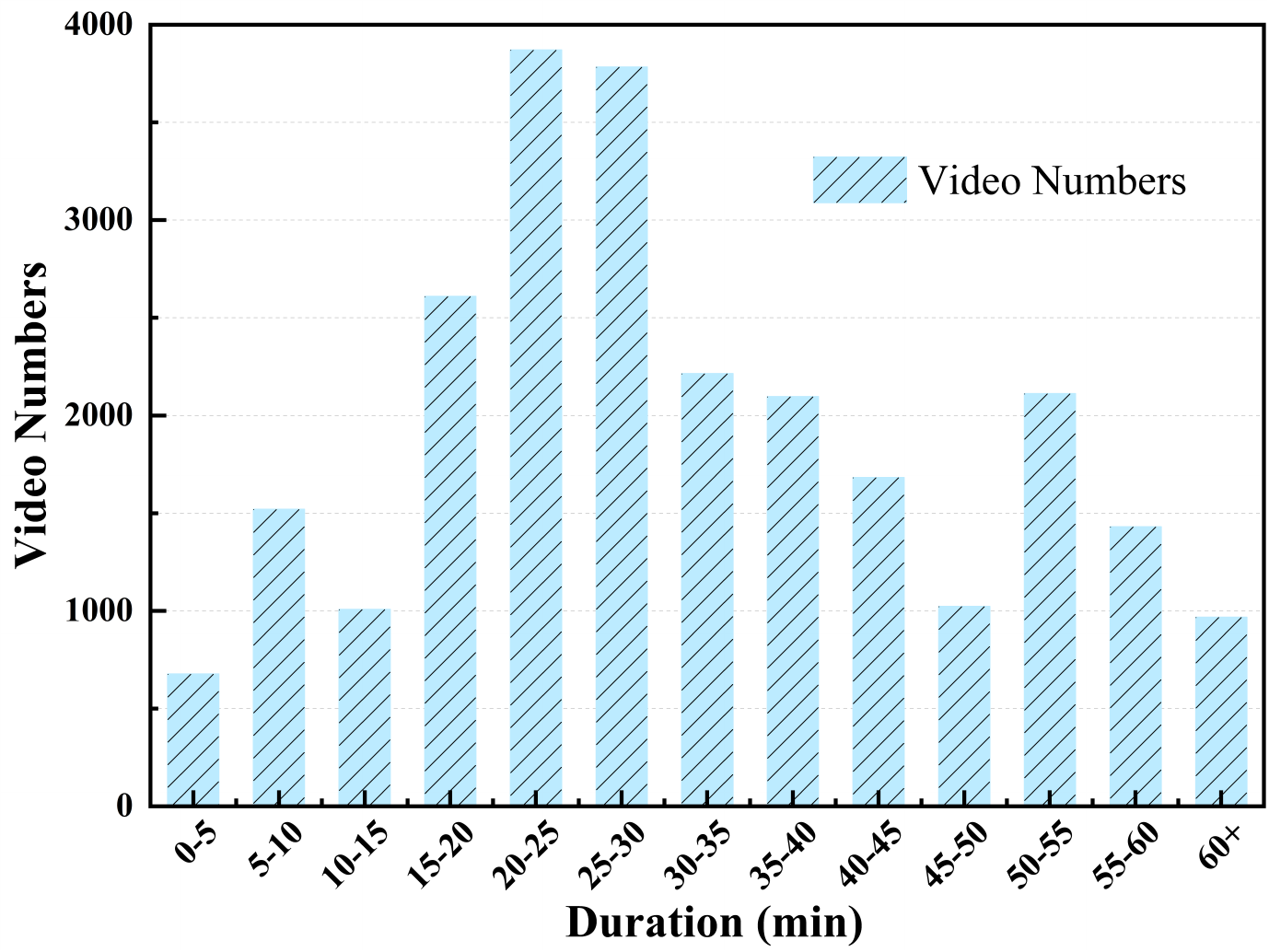}
  \caption{\textbf{Video duration stats.} We analysis the duration distribution of the filtered videos and found that over 60\% were concentrated between 15 and 40 minutes.}
  \label{figure2}
  \vspace{-5.5mm}
\end{figure}
\newline
\textbf{Video Filtering.} During video collection, we set a resolution of 720p or higher and a duration of 1 minute to 2 hours per video. To ensure a balanced representation across categories, our video-collecting team collect no more than 100 clips from each animation. Following MiraData \cite{ju2025miradata}, we filter the videos based on aesthetic quality and NSFW content. Specifically, we first sample frame sequences at a rate of 8 frames per minute for each video, then assess aesthetic quality using the Laion-Aes aesthetic score predictor ~\cite{rombach2022high} and perform NSFW checks with the Stable Diffusion Safety Checker ~\cite{rombach2022high}. After filtering, we retain approximately 25,000 videos, totaling around 11,300 hours, with an average length of about 27.2 minutes per video. Fig.~\ref{figure2} illustrates the duration distribution of the filtered videos.

\begin{figure*}
    \centering
    \includegraphics[width=\linewidth]{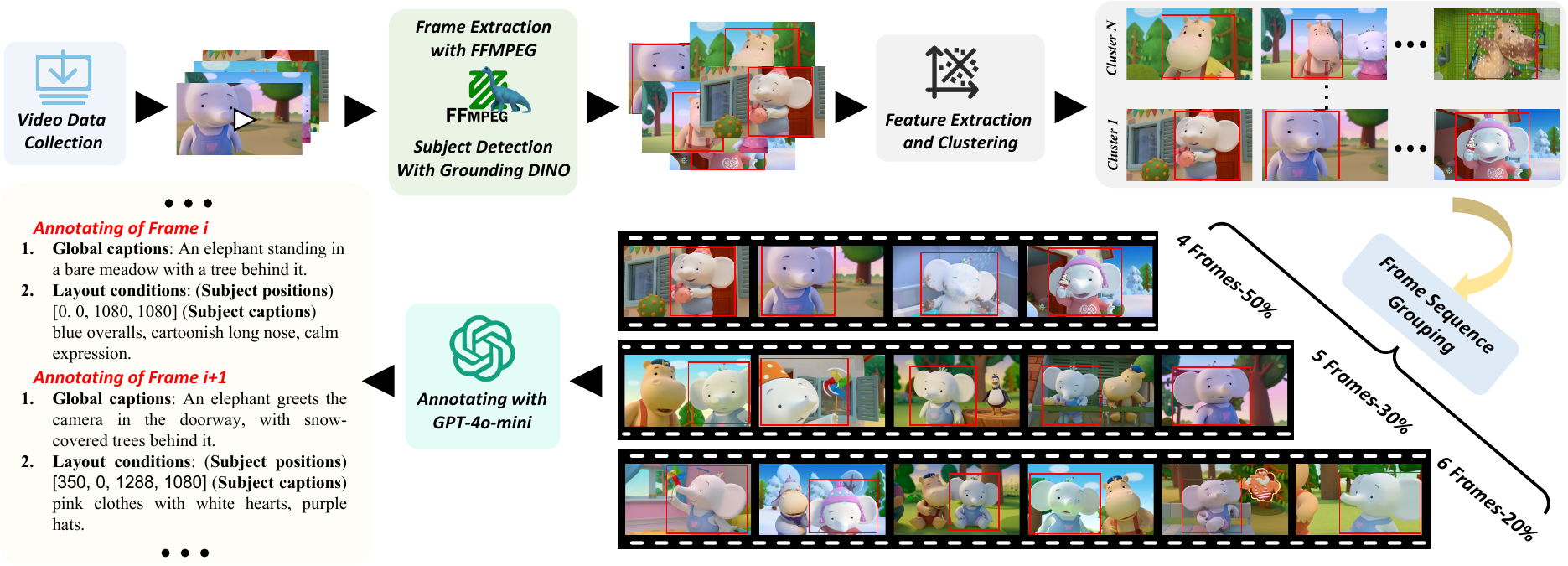}
    \vspace{-6mm}
    \captionof{figure}{
    \textbf{Our data pipeline.} The complete data construction process of the Lay2Story-1M.
    } \vspace{-5mm}
    \label{figure3}
\end{figure*}%

\subsection{Frame Sequence Construction}
\label{Frame Sequence Construction}

As illustrated in Fig.~\ref{figure3}, we develop a robust data processing pipeline to generate frame sequences that meet our training and testing requirements. The pipeline consists of the following key steps:

(1) \textbf{Frame Extraction.} For each input video, we use FFmpeg ~\cite{tomar2006converting} to sample frames at 0.25 FPS. A higher sampling rate may result in excessive frame redundancy, while a lower rate could lead to significant data loss.

(2) \textbf{Subject Detection.} We apply GroundingDINO-B ~\cite{liu2024grounding} to detect subjects in the sampled frames, retaining only the bounding box with the highest detection score to represent the subject’s position.

(3) \textbf{Feature Extraction and Clustering.} Using CLIP-L ~\cite{radford2021learning}, we extract visual features of the main subject area in each frame. We then perform K-means clustering ~\cite{ahmed2020k} to group similar frames. To balance clustering efficiency and effectiveness, we cluster every 150 frames, retaining smaller sets when necessary. For sets of 150 frames, we use 12 cluster centers, whereas for sets with fewer than 150 frames, we use 6 cluster centers.

(4) \textbf{Grouping.} After clustering, we organize frames into sets of 4, 5, and 6 according to a predefined probability distribution, enriching our training and testing data.

(5) \textbf{Annotation.} For each grouped frame sequence, we utilize GPT-4o mini ~\cite{hurst2024gpt} to generate structured annotations. Initially, we prompt the model with multiple frames to produce a global description in the format of ``identity prompt + frame prompt''. Next, for each individual frame, we extract the subject’s image based on its location coordinates and prompt the model to provide a detailed description, capturing key aspects such as appearance, clothing, expression, posture, and other relevant details.

This pipeline ensures high-quality dataset generation while maintaining consistency and diversity in training and evaluation. Using this pipeline, we process the filtered videos from Sec.~\ref{Data Collection and Filtering}, resulting in a dataset of approximately 1.02 million images, named Lay2Story-1M. In this dataset, frame sequences of lengths 4, 5, and 6 account for 50\%, 30\%, and 20\% of the total number of frames, respectively.

\begin{table}[h]
\caption{\textbf{Lay2Story-Bench vs. ConsiStory.} Compared to the ConsiStory benchmark, Lay2Story-Bench offers more prompt sets, higher-quality (HQ) frames, and detailed subject annotations.}
\resizebox{\linewidth}{!}{
\begin{tabular}{ccccc}
\hline
Benchmarks      & \multicolumn{1}{c}{Prompt} & \multicolumn{1}{c}{Prompts} & \multicolumn{1}{c}{HQ Original} & \multicolumn{1}{c}{Detailed Annotations} \\
& \multicolumn{1}{c}{Sets Num} & \multicolumn{1}{c}{Num}  & \multicolumn{1}{c}{Frames} & \multicolumn{1}{c}{of Subjects} \\
\hline

ConsiStory      & 100             & 500         & no                 & no                               \\
\textbf{Lay2Story-Bench} & 655             & 3,000       & yes                & yes                              \\ \hline
\end{tabular}}
\vspace{-4mm}
\label{table2}
\end{table}

\subsection{Lay2Story-Bench}
\label{Lay2Story-Bench}
We curate Lay2Story-Bench by selecting 3,000 samples from Lay2Story-1M, ensuring that the proportions of frame sequences with lengths 4, 5, and 6 align with those in the training set. This results in a test set containing 375 sequences of length 4, 180 of length 5, and 100 of length 6. To enhance the visual quality of the final test results, we select sequences from the top 10\% of videos ranked by aesthetics (as described in Sec.~\ref{Data Collection and Filtering}).

Furthermore, we manually maintain the diversity and balance of the test dataset by limiting the selection to no more than eight frame sequence sets per video category. To ensure fairness, we record the video IDs of selected sequences and guarantee that no other sequences from the same videos appear in the training set.

Overall, as shown in Tab.~\ref{table2}, compared to the previously most commonly used benchmark, ConsiStory ~\cite{tewel2024training}, Lay2Story-Bench provides a larger set of prompts, greater prompt diversity, and high-resolution original frames as ground truth (GT), along with detailed subject annotations. Some examples from Lay2Story-Bench can be found in the \textcolor{blue}{\textbf{Supplementary Material B.2.}}


\begin{figure*}
    \centering
    \includegraphics[width=\linewidth]{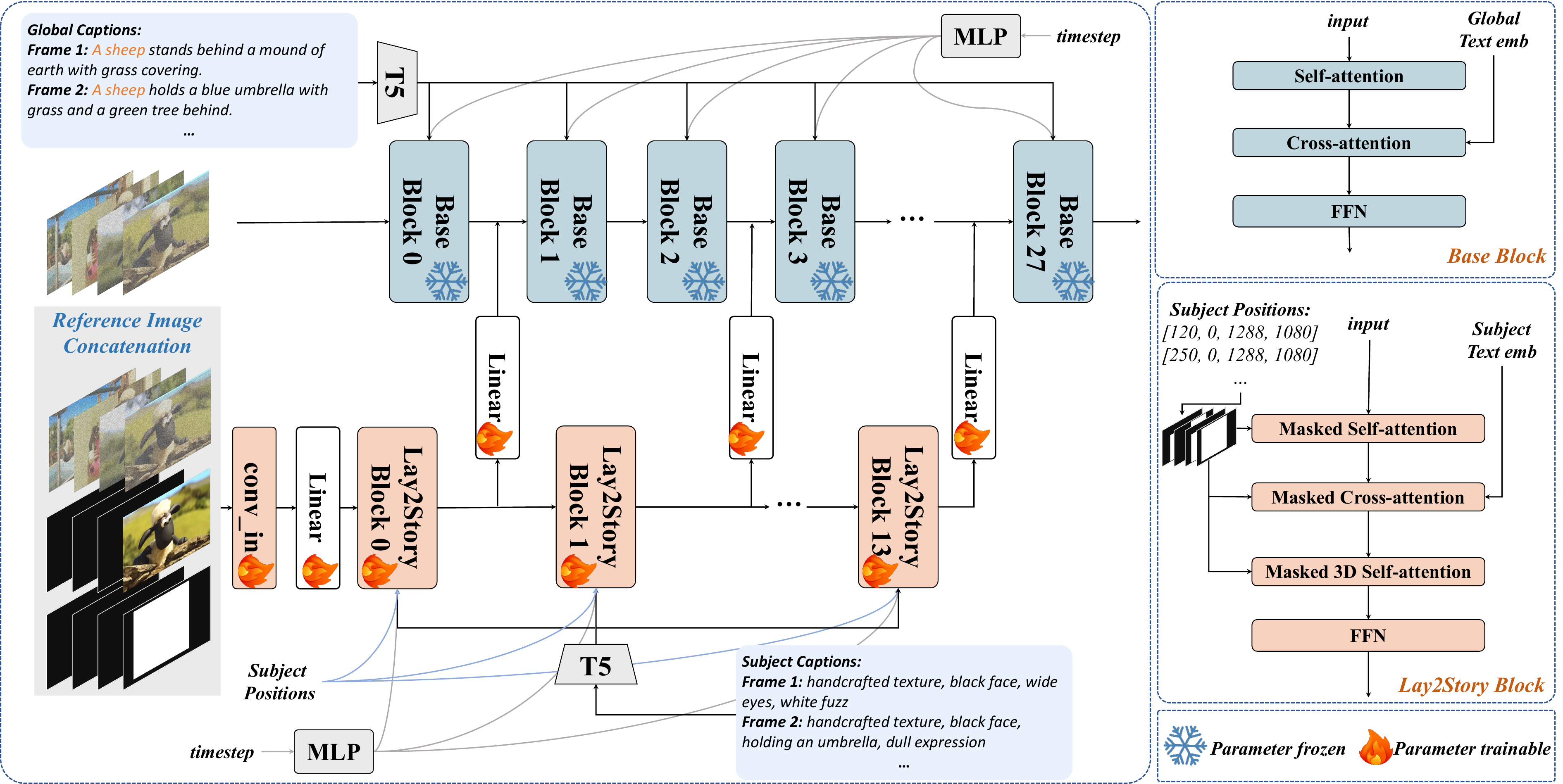}
    \vspace{-6mm}
    \captionof{figure}{
    \textbf{The framework of Lay2Story.} Our Lay2Story consists of two main branches: the global branch and the subject branch. The global branch, inspired by PixArt-$\alpha$, is composed of self-attention and cross-attention, focusing on the overall image quality, guided by global captions. The subject branch, composed of masked self-attention, masked cross-attention, and masked 3D self-attention, ensures the subject's appearance consistency across different images, guided by subject captions and the subject mask. The input to the subject branch consists of the noise latent, the latent of the reference image, and the given image mask of the reference image.
    } 
    \label{figure4}
    \vspace{-5mm}
\end{figure*}%

\section{Method}
\subsection{Preliminary}
Please refer to the \textcolor{blue}{\textbf{Supplementary Material C}}.

\subsection{Lay2Story}
In the Layout-Togglable Storytelling task, the model needs to precisely control the position and appearance of the subject in the image based on the subject's location and descriptive caption. To address this, we propose the Lay2Story, as shown in Fig.~\ref{figure4}. It is a story generation model that requires a reference image $\mathcal{I}_{ref}$ along with its corresponding bounding boxes $\mathcal{B}_{ref}$ and the subject's bounding boxes $\mathcal{B}$ in each image of the story. The bounding boxes $\mathcal{B}_{ref}$ and $\mathcal{B}$ are processed into mask $\mathcal{M}_{ref}$ and $\mathcal{M}$ for further computation. Lay2Story consists of two main branches: the \textbf{global branch} and the \textbf{subject branch}. The input to the global branch is the noise latent $\mathcal{Z}$, guided by global captions, which focus on the overall quality of the image. As for the input to the subject branch, it consists of the noise latent $\mathcal{Z}$, the latent of the reference image $\mathcal{I}_{ref}$, and the given image mask $\mathcal{M}_{ref}$ of the given image. In the subject branch, the model uses subject captions as conditions and utilizes the subject's position $\mathcal{M}$ to constrain the range of attention calculations, ensuring that attention is computed only within the area containing the subject. Additionally, a skip connection design is employed, where the output of each block is returned to the global branch, facilitating a better fusion of global and local information. To ensure consistency of the subject's appearance across different images of a story, the subject branch introduces a 3D attention mechanism and uses subject captions as guidance.

\subsubsection{Global Branch}
We use the PixArt-$\alpha$ model, fine-tuned on the Lay2Story training set, as our global branch. Each Transformer block in PixArt-$\alpha$ consists of three core components: AdaLN-single, self-attention, and cross-attention.

\noindent\textbf{AdaLN-single.} In AdaLN-single, a global set of shift and scale parameters is computed using the time embedding in the first block. These parameters are then shared across all blocks. For each block, the parameters are adapted using a layer-specific trainable embedding, allowing the model to adjust the shift and scale independently in each block. Specifically, given the current timestep $t$ as input, it is mapped through multiple MLP layers to obtain six parameters [$\beta_1^{(i)}, \beta_2^{(i)}, \gamma_1^{(i)}, \gamma_2^{(i)}, \alpha_1^{(i)}, \alpha_2^{(i)}$], which adjust the scale and shift parameters in different blocks.

\noindent\textbf{Self-attention Layer.} The self-attention mechanism in the DiTs plays a key role in capturing dependencies between different parts of the input data. For the $n$-th block, the self-attention layer takes the noise latent $\mathcal{Z}^{n-1}$ output from the ($n-1$)-th block as its input and uses it as the queries (Q), keys (K), and values (V).

\noindent\textbf{Cross-attention Layer.} A multi-head cross-attention layer is inserted between the self-attention and feed-forward layers. This enables the model to interact flexibly with the global text embeddings $TM_{global}$, which are obtained by applying the language model T5. This can be written as $TM_{global}=F_{T5}(T_{global})$. For the cross-attention layer, we use the noise latent as Query (Q) and the global text embeddings $TM_{global}$ as Key (K) and Value (V).

\subsubsection{Subject Branch}
The design of the subject branch in Lay2Story is intended to enable more controllable story generation. The design of this branch is inspired by ControlNet. To save computational resources, we introduce an output from the Subject branch block after every two global branch blocks. To improve training efficiency, we partially reuse parameters from the global branch. The guiding conditions for the subject branch include the subject's masks $\mathcal{M}$ and its detailed description $\mathcal{T}_{subject}$. The core of this branch consists of four key components: AdaLN-single (which follows the settings of PixArt-$\alpha$), masked self-attention, masked cross-attention, and masked 3D self-attention. The computation of the masked attention $MA(\cdot,\cdot,\cdot,\cdot)$ can be elegantly unified as follows:

\begin{equation}
\text{MA}(Q, K, V, M) = \text{SF}\left( \frac{QK^T}{\sqrt{d_k}} + M \right) V
\end{equation}
where $M$ represents the mask, which has the same dimensions as $QK^T$, with the background values set to a large number, SF$(\cdot)$ denotes the softmax function.

\noindent\textbf{Reference Image Concatenation.}
As for the input, we first pass the reference image $\mathcal{I}$ through the VAE to obtain a four-channel feature map $\mathcal{F}_{rep} \in \mathbb{R}^{b \times h \times w \times 4}$. Based on the subject’s bounding box $\mathcal{B}_{ref}$ information in the frame, we generate a mask $\mathcal{M}_{ref} \in \mathbb{R}^{b \times h \times w \times 1}$ for the reference image. We first pad $\mathcal{F}_{rep}$ and $\mathcal{M}_{ref}$ with zeros, expanding them to match the same number of $f$ as the noise latent $\mathcal{Z} \in \mathbb{R}^{b \times f \times h \times w \times 4}$. Then, we concatenate the image features $\mathcal{F}_{rep}$, mask $\mathcal{M}_{ref}$, and noise latent$\mathcal{Z}$ along the channel dimension to obtain a 9-channel output $\mathcal{Z}_{con}\in \mathbb{R}^{b \times f \times h \times w \times 9}$. Finally, a convolutional layer reduces the 9 channels to 4 channels, producing the subject noise latent $\mathcal{Z}_{sub} \in \mathbb{R}^{b \times f \times h \times w \times 4}$ as input of the subject branch.

\noindent\textbf{Masked Self-attention Layer.} 
To focus more on the spatial context of the subject region, we employ masked self-attention. Given the bounding boxes $\mathcal{B}$ of the subjects across frames, we generate the mask $\mathcal{M}_{s} \in \mathbb{R}^{b \times f \times (h w)\times (h w) }$ to serve as an input mask during the self-attention mechanism. For the $m$-th block, the masked self-attention layer takes the noise latent $\mathcal{Z}_{sub}^{m-1}$ output from the ($m-1$)-th block as its input and uses it as the queries (Q), keys (K), and values (V), with $M$ serving as the mask.

To accommodate scenarios where the bounding boxes are not provided by the user, we initialize 25\% of the mask $\mathcal{M}_{s}$ to be fully set to valid values, effectively labeling the entire region as subject areas during training. This strategy encourages the model to prioritize subject regions while maintaining spatial awareness and robustness in scenarios with missing annotations. 

\noindent\textbf{Masked Cross-attention Layer.} 
To incorporate more fine-grained and detailed descriptions of the subject’s appearance, we introduce masked cross-attention layer. We first use T5 to obtain the subject text embeddings $TM_{subject}$ from the subject caption $T_{subject}$. This can be written as $TM_{subject}=F_{T5}(T_{subject})$ In masked cross-attention, the mask $\mathcal{M}_{c} \in \mathbb{R}^{b \times f \times (h w)\times l }$ is applied, where $l$ is the length of caption $T_{subject}$. For masked cross-attention layer, we use the subject noise latent as Query (Q), the subject text embeddings $TM_{global}$ as Key (K) and Value (V), and the mask $\mathcal{M}_{c}$ as the attention mask. This design aims to enhance the model’s ability to capture detailed subject attributes while maintaining spatial coherence. 

Additionally, to accommodate scenarios where the subject captions are not explicitly provided by the user, we introduce a training strategy where, with a 25\% probability, the subject caption $T_{subject}$ is randomly replaced with the corresponding global caption $T_{global}$. 

\noindent\textbf{Masked 3D Self-attention Layer.} To ensure subject consistency across different images, we draw from video generation tasks and design a masked 3D self-attention mechanism. Specifically, we first reshape subject noise latent $\mathcal{Z}_{sub} \in \mathbb{R}^{b \times f \times (h w) \times c} $ to $\mathbb{R}^{b \times (f h w) \times c}$, enabling cross-frame information propagation. Similarly, an attention mask $\mathcal{M}_t \in \mathbb{R}^{b \times (f h w)\times (f h w) }$ is applied to constrain the model’s focus within the subject positions, ensuring consistency in subject representation. For masked 3D self-attention layer, we use the subject noise latent $\mathcal{Z}_{sub}$ as Query (Q), Key (K) and Value (V) and $\mathcal{M}_t$ as mask. This layer effectively ensures subject consistency within the story generation model.

\noindent\textbf{Information Propagation from the Subject Branch.}
We adopt the same approach as ControlNet to propagate the updated features from the subject branch to the global branch. Due to the skip connection structure we use, after every two Base Blocks in the global branch, the output will simultaneously receive an output from the subject branch, which has been processed by a zero linear. This can be expressed as:

\begin{equation}
\mathcal{Z}^{n} = \mathcal{Z}^{n} + F_m(\mathcal{Z}^{m}_{sub})
\end{equation}
where $F_m$ denotes the linear layer with zero initialization, and $n$ is twice the value of $m$.

\begin{figure*}
    \centering
    \includegraphics[width=\linewidth]{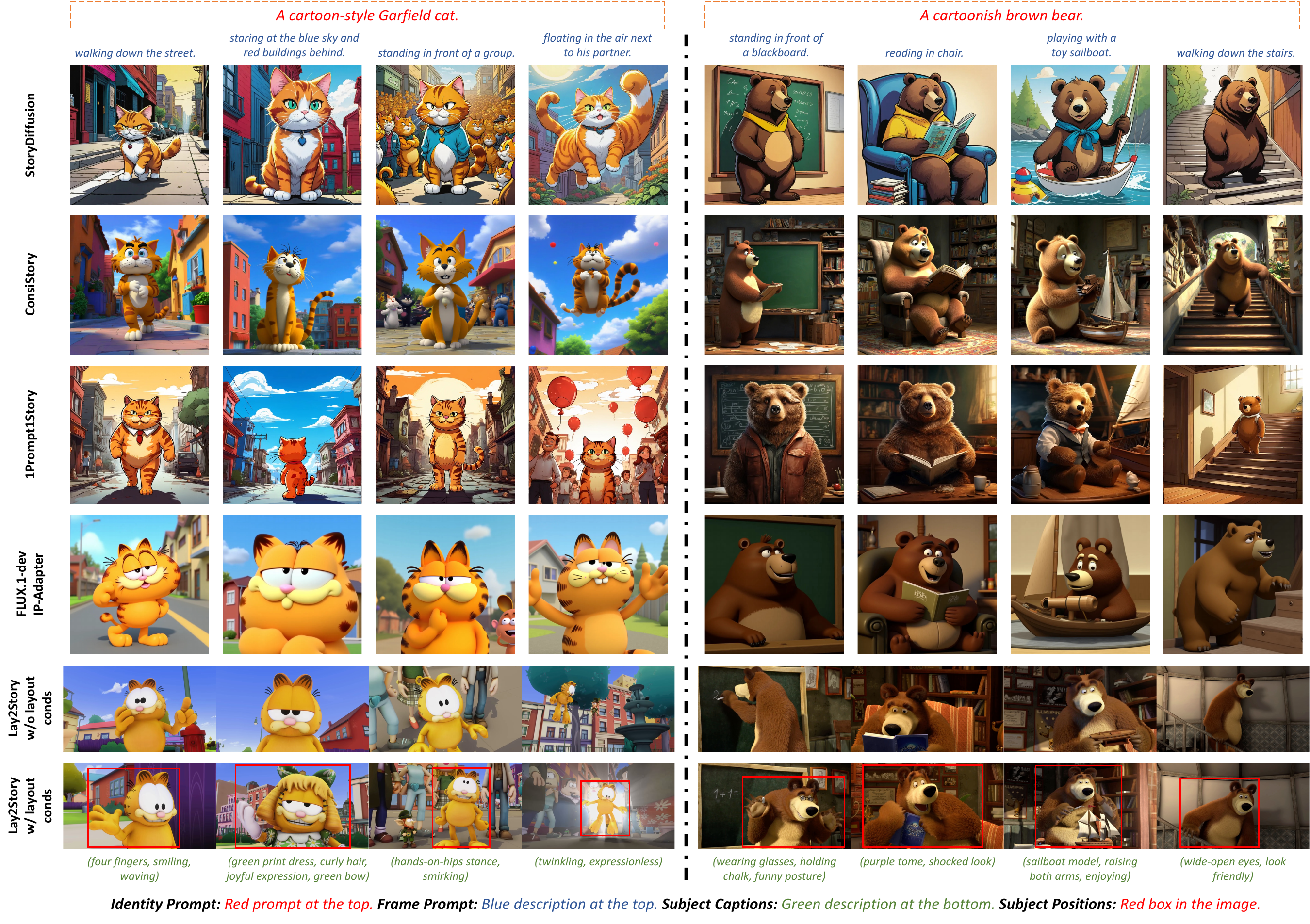}
    \vspace{-6mm}
    \captionof{figure}{
    \textbf{Qualitative results.} We compare our Lay2Story with StoryDiffusion, ConsiStory, 1Prompt1Story, and FLUX.1-dev IP-Adapter.
    Our method shows the impact of layout conditions (last two rows). Without them, it uses identity and frame prompts; otherwise, it adds subject captions and positions for finer control. Lay2Story generates images at 720 $\times$ 1280 resolution, while others use 1024 $\times$ 1024.
    } \vspace{-3mm}
    \label{figure5}
\end{figure*}%

\vspace{-1mm}
\section{Experiments}

\subsection{Implementation Details}
\textbf{Training and Inference Settings.} Please refer to the \textcolor{blue}{\textbf{Supplementary Material D}}.
\newline
\textbf{Baselines and Benchmarks.} We compare our approach with the following methods for story generation: BLIP-Diffusion ~\cite{li2023blip}, StoryGen ~\cite{liu2024intelligent}, ConsiStory \cite{tewel2024training}, StoryDiffusion ~\cite{zhou2025storydiffusion}, 1Prompt1Story ~\cite{liu2025one}, and FLUX.1-dev IP-Adapter ~\cite{flux-ipa}. We adopt the default configuration specified in their paper or open-source implementation. 

In the evaluation phase, we conduct both qualitative and quantitative comparisons of our method with the aforementioned approaches using the Lay2Story-Bench. In the qualitative comparison, we present the inference results of all methods consistently, using a prompt set of 4 frames. We should highlight that, due to the togglable nature of our method in handling layout conditions (including subject positions and captions) as input, we present the generated results both with and without the inclusion of these inputs, respectively. In the quantitative comparison, all methods are evaluated on the full prompt set of Lay2Story-Bench. Similar to the qualitative comparison, we present the generated results with and without layout conditions inputs.
\newline
\textbf{Evaluation Methods.} Building on the previous methods ~\cite{liu2025one, yang2024seed}, we employ DreamSim ~\cite{fu2023dreamsim} and CLIP-I ~\cite{hessel2021clipscore} to assess subject consistency. To ensure the similarity measurement focuses solely on the subject's identity, we follow the approach outlined in DreamSim ~\cite{fu2023dreamsim}, using CarveKit ~\cite{githubrepo_carvekit} to remove the image background and replace it with random noise. We also utilize the FID ~\cite{heusel2017gans} metrics to assess the quality of the generated images. Recall@1 measures top-1 text-to-image matching accuracy, while human preference reflects averaged binary ratings from three annotators.

\begin{table*}[!htb]
    \centering
    \footnotesize
    \renewcommand{\arraystretch}{1.5}
\renewcommand{\tabcolsep}{5pt}
\caption{\textbf{Quantitative results.} Quantitative evaluation on the Lay2Story Bench comparing Lay2Story with baseline methods. The best and second-best metrics are highlighted in \textbf{bold} and \underline{underlined}, respectively. FLUX.1-dev IPA refers to FLUX.1-dev IP-Adapter. Human-Pre stands for human preference. Lay2Story w/o lc denotes no injected layout conditions, while w/ lc indicates injected layout conditions.}

\label{table3}
\begin{tabular}{cc|ccccccc}
\toprule[1.2pt]
\multicolumn{1}{c}{Architecture} & Method                                                             & \gb{DreamSim ($\downarrow$)}         & \gb{CLIP-I ($\uparrow$)}    & \gb{FID ($\downarrow$)}     & \gb{Recall@1 ($\uparrow$)} & \gb{Human-Pre ($\uparrow$)}              & \gb{Steps} & \gb{Inference Time(s) ($\downarrow$)} 
\\ \toprule[1.2pt]
\multirow{5}{*}{U-Net}
& BLIP-Diffusion ~\cite{li2023blip}          
& 0.3117          & 0.7527    & 113.72       & 0.3675          & 0.3956                   & 26    & \textbf{4.32}               \\
& StoryGen ~\cite{liu2024intelligent}              
& 0.3358          & 0.7423   & 96.38        & 0.4933          & 0.4768                    & 50    & 19.15              \\
& ConsiStory ~\cite{tewel2024training} 
& 0.2615          & 0.8084    & 75.54       & 0.5672          & 0.6620                 & 50    & 38.51              \\
& StoryDiffusion ~\cite{zhou2025storydiffusion}               
& 0.2936          & 0.8016    & 82.17        & 0.5320          & 0.6437                  & 50    & 24.36              \\
& 1Prompt1Story ~\cite{liu2025one}
& 0.2429          & 0.8461    & 66.79       & 0.5583          & 0.6742                  & 50    & 20.69              \\ \toprule[1.2pt]

\multirow{3}{*}{DiT}
& FLUX.1-dev IPA ~\cite{flux-ipa}
& \underline{0.1533}    & 0.9138   & \underline{33.18}       & \underline{0.6482}    &  0.7059           & 25    & 61.38              \\

& Lay2Story w/o lc                                               
& 0.1602          & \underline{0.9214}  & 35.82  &  0.6376        &  \underline{0.7123}         & 25    & \underline{13.63}              \\      
& Lay2Story w/ lc                                                 
& \textbf{0.1324} & \textbf{0.9299} & \textbf{26.71} & \textbf{0.7012}  & \textbf{0.7561}  & 25    & 14.02             \\   \toprule[1.2pt]
\end{tabular}
\vspace{-3mm}
\end{table*}

\subsection{Qualitative Results}
As shown in Fig.~\ref{figure5}, we qualitatively compare our Lay2Story to StoryDiffusion, ConsiStory, 1Prompt1Story, and FLUX.1-dev IP-Adapter. Previous studies have struggled to maintain subject consistency in image sequences (e.g., the brown bear in StoryDiffusion and the Garfield cat in ConsiStory). Additionally, they exhibit semantic correlation errors (e.g., frame 7 in 1Prompt1Story and frame 4 in FLUX.1-dev IP-Adapter) and distortions in aesthetic quality (e.g., frame 4 in StoryDiffusion and frame 4 in ConsiStory). In a comprehensive comparison, our Lay2Story approach outperforms baseline methods in consistency, semantic relevance, and aesthetic quality. We also present the results of Lay2Story with and without layout conditions (the last two rows). With layout conditions (including subject captions and positions), Lay2Story allows finer control over the subject (e.g., in the last row, Garfield wears a green dress in frame 2 and the bear wears glasses in frame 5).

\subsection{Quantitative Results}
As shown in Tab.~\ref{table3}, we quantitatively compare Lay2Story (with and without layout conditions input) to the baseline methods ~\cite{li2023blip, liu2024intelligent, tewel2024training, zhou2025storydiffusion, liu2025one, flux-ipa}. For subject consistency metrics (DreamSim and CLIP-I), Lay2Story outperforms all other methods when layout conditions are provided, with CLIP-I surpassing the second-best by approximately 1.6 percentage points and DreamSim by about 2 percentage points. Regarding semantic relevance (Recall@1), Lay2Story also outperforms all other methods with layout conditions input, exceeding the second-best by approximately 2 percentage points. For aesthetic quality (FID), Lay2Story significantly surpasses all other methods under layout conditions input, outperforming the second-best by approximately 6.4 percentage points. Moreover, Lay2Story demonstrates competitive performance even without layout conditions input, ranking second in CLIP-I and third in Recall@1, DreamSim, and FID. Additionally, we compare the inference time of various methods and find that Lay2Story exhibits minimal difference (0.4 seconds) when layout conditions are not provided. It is only slightly slower than BLIP-Diffusion while outperforming all other methods.

\begin{table}
\centering
\footnotesize
\caption{\textbf{Ablation of key components.} We evaluate the impact of several key component in Lay2Story, including the subject branch, reference image concatenation, and masked 3D self-attention.}
\centering
\small
\setlength{\tabcolsep}{4pt} 
\begin{tabular}{@{}cccc@{}}
\toprule 
\textbf{Setting} & \textbf{FID ($\downarrow$)} & \textbf{Recall@1 ($\uparrow$)} & \textbf{Human-Pre ($\uparrow$)} \\
\midrule 
w/o subject branch & 110.58 & 0.1733 & 0.1928 \\
w/o reference image & 50.27 & 0.3981 & 0.4781 \\
w/o masked 3d sa & 66.14 & 0.3274 & 0.3982 \\
Lay2Story & 26.71 & 0.7012 & 0.7561 \\
\bottomrule 
\end{tabular}
\vspace{-6.5mm}
\label{table4}
\end{table}

\subsection{Ablation Study}
\textbf{Ablation of Key Components in Lay2Story.}
As shown in Tab.~\ref{table4}, we evaluate the impact of key components in Lay2Story, including the subject branch, reference image concatenation, and masked 3D self-attention layer. The effectiveness of key components in Lay2Story is evaluated through comprehensive experiments from three perspectives—FID, Recall@1, and human preference.

\begin{figure}
  \centering
    \includegraphics[width=1.0\linewidth]{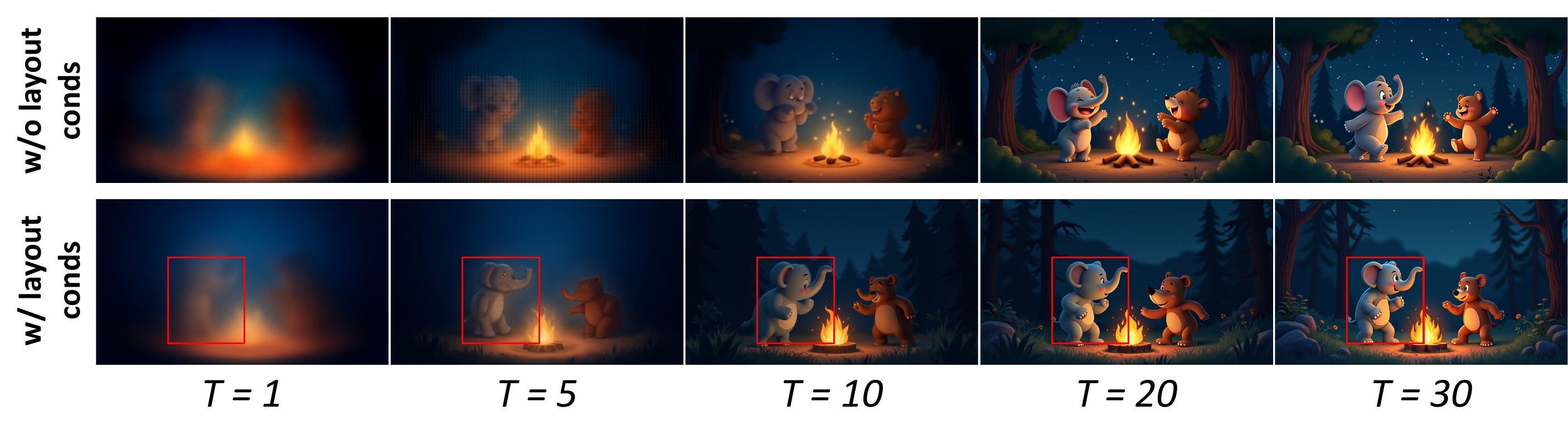}
    \vspace{-6mm}
  \caption{\textbf{Ablation of layout conditions.} We run the same Lay2Story model with and without layout conditions, recording images at denoising steps 1, 5, 10, 20, and 30.}
  \label{figure6}
  \vspace{-6.5mm}
\end{figure}

\textbf{Ablation of Layout Conditions Input.}
As shown in Fig.~\ref{figure6}, we evaluate the performance of the Lay2Story model by varying the denoising steps with and without layout conditions as input. The test prompt is, ``An elephant and a little bear dance around a campfire in the woods at night.'' It is clear that the model performs better with layout conditions, especially in the early denoising stages. For example, the small elephant at T=5 is more defined with layout conditions. In the later stages (T=30), the image quality with layout conditions also surpasses that without.

\section{Conclusion}
In this paper, we introduce an advanced version of the storytelling task: Layout-Togglable Storytelling, which allows for precise subject control by incorporating layout conditions, with users having the option to choose whether to include them. We present Lay2Story-1M, a dataset containing over 1 million high-quality images at 720p resolution or higher, along with detailed subject annotations. Building on Lay2Story-1M, we introduce Lay2Story-Bench, a benchmark consisting of 3,000 prompts and corresponding high-quality images to evaluate our model and comparison methods. We also propose Lay2Story, a training-based approach built on the DiTs architecture. Through extensive comparisons with existing storytelling methods, we demonstrate that Lay2Story outperforms relevant approaches in terms of consistency, semantic correlation, and aesthetic quality.
\renewcommand{\thesection}{\Alph{section}}
\setcounter{section}{0} 
\section{Related Work}
\subsection{Consistent Text-to-image Generation}
Consistent image generation methods can be categorized into high-level semantic consistency, facial consistency, style consistency~\cite{bi2024using}, and object consistency~\cite{zhang2024survey}.
High-level semantic consistency methods \cite{huang2024reversion,motamed2023lego,huang2024learning,zheng2023layoutdiffusion}, such as ReVersion \cite{huang2024reversion}, achieve consistency by inverting object relations and utilizing a contrastive loss to guide the optimization of token embeddings toward specific clusters of Part-of-Speech tags, such as prepositions, nouns, and verbs. 
Facial consistency methods \cite{zhou2023enhancing,valevski2023face0,yan2023facestudio,li2024photomaker}, such as PhotoMaker \cite{li2024photomaker}, construct high-quality datasets through meticulous data collection and filtering pipelines, employing a two-layer MLP to fuse ID features and class embeddings for comprehensive human portrait representation. 
Style consistency methods \cite{sohn2023styledrop,zhang2024generative,hertz2024style,wang2023styleadapter}, such as StyleAdapter \cite{wang2023styleadapter}, introduce a specialized embedding module to extract and integrate global features from multiple style references and employ a dual-path cross-attention mechanism within the learning framework. 
Object consistency methods \cite{gal2022image,voronov2023loss,voynov2023p+} include approaches like IP-Adapter \cite{ye2023ip}, which trains a lightweight decoupled cross-attention module where image and text features are processed separately with query features; DreamBooth \cite{ruiz2023dreambooth}, which proposes using a unique modifier with a rare token to represent the subject of interest and fine-tuning all parameters of the diffusion model; and UMM-Diffusion \cite{ma2023unified}, which designs a multi-modal encoder to generate fused features based on the reference image and text prompt. Storytelling task can essentially be categorized as an object consistency image generation task, aiming to achieve consistent visual narratives through cross-modal fusion \cite{zhang2024survey}.

\subsection{Layout-to-image Generation}
Layout-controllable image generation aims to apply layout control to place subjects in user-defined positions within an image, which has become an active research area \cite{zhao2019image,xue2023freestyle,he2021context,ma2024hico}.
SimM \cite{gong2024check} is a training-free system that corrects layout errors during inference by analyzing prompts, detecting inconsistencies, and adjusting activations.
ReCo \cite{yang2023reco} introduces a unified token vocabulary containing both text and positional tokens for precise, open-ended regional control.
InteractDiffusion \cite{hoe2024interactdiffusion} enhances T2I diffusion models by incorporating Human-Object Interaction (HOI) information through tokenized embeddings and a self-attention layer, enabling better control of interactions and locations in generated images.
CreatiLayout \cite{zhang2024creatilayout} introduces a Siamese architecture to decouple image-layout interactions in MM-DiT, treating layout as an independent modality and integrating it with text and image features while leveraging a large-scale dataset for training and evaluation.
Combining the Layout-to-Image task with the Storytelling task is both innovative and valuable.

\subsection{Storytelling Generation}
Generating a sequence of frames with a consistent subject from a given script, known as storytelling, is a rapidly evolving field. Current methods are generally categorized into two types: training-free and training-based. Training-free methods, such as StoryDiffusion ~\cite{zhou2025storydiffusion}, utilize consistent self-attention computation based on the SD1.5 ~\cite{rombach2022high} model to maintain subject consistency throughout the story sequence. ConsiStory ~\cite{tewel2024training} achieves subject consistency by sharing the internal activations of the pre-trained diffusion model. 1Prompt1Story ~\cite{liu2025one} takes advantage of the inherent context consistency of language models, using a single prompt to generate a cohesive narrative across the story sequence. Training-based methods, such as Seed-Story ~\cite{yang2024seed}, employ the Multimodal Large Language Model (MLLM) to predict text and visual tokens, followed by a visual de-tokenizer to ensure subject consistency across the image sequence. FLUX.1-dev IP-Adapter ~\cite{flux-ipa} builds upon the robust image generation model FLUX ~\cite{githubrepo_flux}, training an adapter to integrate reference image features, enabling FLUX to generate images while leveraging the reference image conditions to maintain consistency.

In this paper, we propose a training-based method, Lay2Story, which not only keeps the subject consistent but also enables more refined control over the subject by injecting layout conditions into the model, including its position, appearance, clothing, expression, posture, and other relevant details. Our model consists of two main branches: the global branch and the subject branch. The global branch takes noise latent as input, guided by global captions, and focuses on generating the overall image content. The subject branch takes as input the noise latent, subject mask, and latent vector of a reference image, guided by subject captions and focuses on maintaining subject consistency while generating the subject's position and detailed attributes. The Lay2Story model, built on Diffusion Transformers (DiTs), is based on the PixArt-$\alpha$ ~\cite{chen2023pixarta} image generation model. Inspired by MM-DiT, Lay2Story employs Masked 3D Self-Attention to enhance subject consistency through inter-frame attention guided by subject masks. Unlike StoryDiffusion, it is trained on consistent sequences; unlike Storynizor, it additionally incorporates subject information for more precise layout control. During training, we first fine-tune the base model with image data from Lay2Story-1M, then freeze the global branches and train the subject branches on a consistent frame sequence. This enables our model to simultaneously achieve consistency, semantic relevance, and aesthetic quality.

\begin{figure*}
    \centering
    \includegraphics[width=\linewidth]{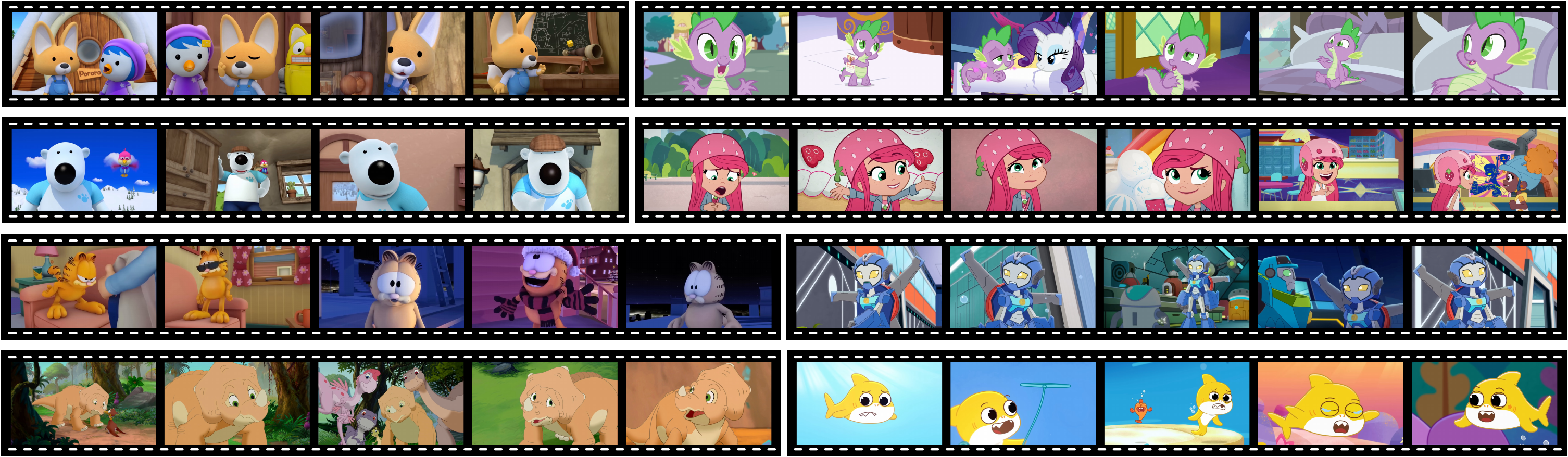}
    \vspace{-6mm}
    \captionof{figure}{
    \textbf{Frame sequence examples.} We present renderings of several frame sequences from Lay2Story-1M.
    } \vspace{-3mm}
    \label{supp_fig1}
\end{figure*}%

\begin{figure*}
    \centering
    \includegraphics[width=\linewidth]{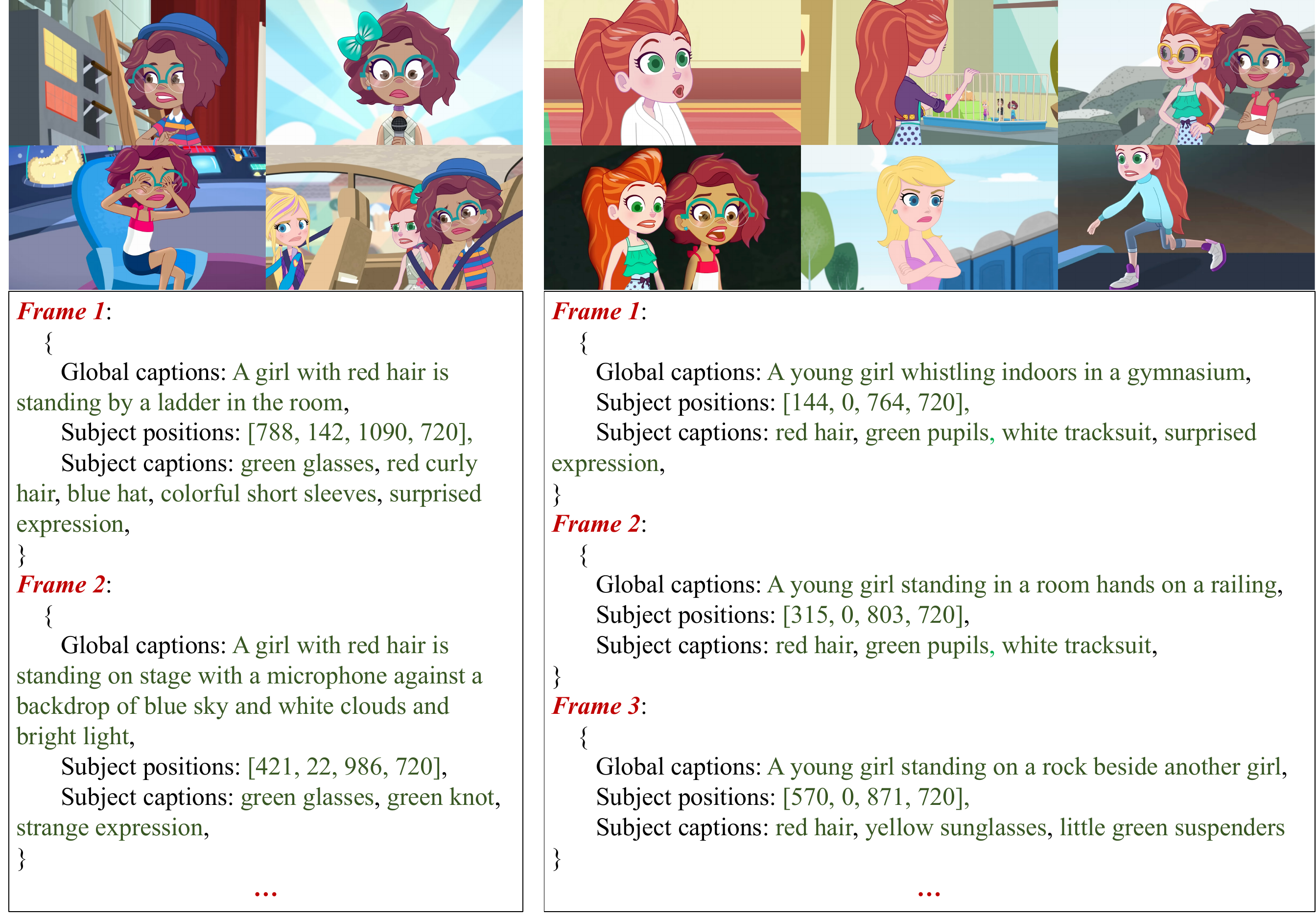}
    \vspace{-6mm}
    \captionof{figure}{
    \textbf{Examples of Lay2Story-Bench.} We present examples from the Lay2Story-Bench benchmark, including the original images and annotations, which consist of global captions, subject positions, and subject captions for each frame.
    } \vspace{-3mm}
    \label{supp_fig2}
\end{figure*}%

\section{Examples of Lay2Story-1M}
\subsection{Frame Sequence Examples}
As shown in Fig.~\ref{supp_fig1}, we provide the image data of frame sequences from the Lay2Story-1M dataset (without showing annotation information such as global captions or layout conditions), with sequence lengths ranging from 4 to 6 frames.

\subsection{Examples of Lay2Story-Bench}
As shown in Fig.~\ref{supp_fig2}, we present examples from Lay2Story-Bench, including raw frame sequence images and their corresponding annotations, which cover global captions, subject positions, and subject captions for each frame.

\section{Preliminary}
\subsection{Latent Diffusion Models}
Latent diffusion models ~\cite{rombach2022high,he2024freeedit,shao2025tr,shao2025context,chen2025ctr,li2023planning} learn a denoising process to simulate the probability distribution within latent space. To reduce the computational load, the image \( x \) is transformed into a latent space feature \( z_0 = E(x) \) using a Variational Autoencoder (VAE) Encoder \( E \) \cite{han2019variational}. During the forward diffusion process, Gaussian noise is iteratively added to \( z_0 \) at timesteps \( t \), resulting in \( z_t \), according to the equation:

\begin{equation}
q(z_t | z_{t-1}) = \mathcal{N}(z_t; \sqrt{1 - \beta_t} z_{t-1}, \beta_t I)
\end{equation}
where \( \beta_t \) represents a sequence schedule. The denoising process is defined as an iterative Markov Chain that denoises the initial Gaussian noise \( z_T \in \mathcal{N}(0, I) \) into the clean latent space \( z_0 \). The denoising function in LDM is typically implemented with U-Net ~\cite{ronneberger2015u} or Transformers ~\cite{vaswani2017attention}, trained by minimizing the mean squared error loss:

\begin{equation}
L = \mathbb{E}_{z_t, c, t, \epsilon \sim \mathcal{N}(0, I)} \left[ \| \epsilon - \epsilon_{\theta}(x_t; c, t) \|^2_2 \right]
\end{equation}
where \( \epsilon_{\theta} \) represents the parameterized network for predicting noise, and \( c \) denotes an optional conditional input. Subsequently, the denoised latent space feature is decoded into image pixels using the VAE Decoder \( D \).
\subsection{Diffusion Transformers}
In the task of consistent image generation, improvements are often made to the U-Net model ~\cite{ronneberger2015u}, with common optimizations including SD1.5 ~\cite{rombach2022high} and SDXL ~\cite{podell2023sdxl}. In recent years, Transformer-based approaches have gradually matured in the field of text-to-image generation, with representative methods such as Stable Diffusion 3 ~\cite{esser2024scaling} and PixArt-$\alpha$ ~\cite{chen2023pixarta}. These methods have demonstrated the significant advantages of Diffusion Transformers in terms of scalability, an area where U-Net falls short. The core module of PixArt-$\alpha$ consists of three parts: first, the linear layers that generate scale shift parameters for output normalization; second, a self-attention mechanism with latent inputs to enhance generation quality; and third, a cross-attention mechanism that takes both latent and text embeddings as inputs, using textual information as a condition to guide the generation process.

\section{Training and Inference Settings}
We adopt a similar approach to PixArt-$\alpha$, using T5 ~\cite{raffel2020exploring} as the text encoder with a fixed token length 120. The training process consists of two stages. In the first stage, we fine-tune the global branch for the text-to-image task, training the model with the AdamW optimizer at a learning rate of 2e-5 and a weight decay of 0.03. The model runs for 5 epochs on the Lay2Story-1M dataset using 16 40GB A100 GPUs. In the second stage, we freeze the global branch and train the subject branch of Lay2Story independently, using the AdamW optimizer with a learning rate of 1e-5 and the same weight decay. This stage lasts for 10 epochs with 32 80GB A100 GPUs. During inference, we follow the configuration of previous studies, using 25 sampling steps and setting the class-free guidance coefficient to 4.5.

\section{Supplementary Analyses and Experiments}

\subsection{Computational Complexity Analysis}
Table \ref{table_com} reports GPU memory usage and inference time~\cite{xu2025dropoutgs,wang2025learning,wang2022spnet}.

\begin{table}[htbp!] 
\caption{\textbf{Computational cost}. All experiments were conducted on an 80GB A100 GPU using FlashAttention at a resolution of 720p.
}
\label{table_com}
\centering
\small
\setlength{\tabcolsep}{4pt} 
\begin{tabular}{@{}ccc@{}}
\toprule 
\textbf{Frame num} & \textbf{Inference time (s)} & \textbf{Memory (MiB)} \\
\midrule 
4 & 14.02 & 29731 \\
8 & 17.70 & 33072 \\
16 & 29.69 & 46320 \\
32 & 78.33 & 62127 \\
\bottomrule 
\end{tabular}
\vspace{-8pt}
\end{table}

\subsection{Multi-Subject Experiments}
Owing to the high cost associated with data collection and training, the current experiments are limited to single-subject narratives. Nonetheless, the proposed pipeline is inherently compatible with multi-subject scenarios, as it preserves all subject bounding boxes during the Grounding DINO detection stage, followed by feature extraction, clustering, and grouping. Multi-subject handling is facilitated by concatenating the positional embeddings of all subjects and conditioning the model on the corresponding textual embeddings, thereby maintaining spatial layout and texture consistency. Comprehensive evaluation under multi-subject settings is left as an avenue for future exploration.

{
    \small
    \bibliographystyle{ieeenat_fullname}
    \bibliography{main}
}

\end{document}